\documentclass{article}

\usepackage{hyperref}

\usepackage[utf8]{inputenc} 
\usepackage[T1]{fontenc}    
\usepackage{hyperref}       
\usepackage{url}            
\usepackage{booktabs}       
\usepackage{nicefrac}       
\usepackage{microtype}      
\usepackage{xcolor}         
\usepackage{comment}
\usepackage{xspace}
\usepackage{multirow}
\usepackage{algorithm}%
\usepackage{algpseudocode}%
\usepackage{amsfonts}
\usepackage{amsmath}
\usepackage{makecell}
\usepackage{graphicx}
\usepackage{subcaption}
\usepackage{amsthm}

\usepackage{bbm}
\usepackage{centernot}
\usepackage{natbib}
\newcommand{\ie}{\textit{i.e., }}

\newcommand{\xhdr}[1]{\vspace{2mm}\noindent{{\bf #1.}}}

\DeclareMathOperator*{\argmin}{arg\,min}
\newcommand*{\defeq}{\stackrel{\text{def}}{=}}

\usepackage[accepted]{icml2021}

\icmltitlerunning{A Tale Of Two Long Tails}

\begin{document}

\twocolumn[
\icmltitle{A Tale Of Two Long Tails}



\icmlsetsymbol{equal}{*}

\begin{icmlauthorlist}
\icmlauthor{Daniel D'souza}{mlc,pq}
\icmlauthor{Zach Nussbaum}{mlc}
\icmlauthor{Chirag Agarwal}{harvard}
\icmlauthor{Sara Hooker}{brain}
\end{icmlauthorlist}

\icmlaffiliation{mlc}{ML Collective}
\icmlaffiliation{harvard}{Harvard University}
\icmlaffiliation{brain}{Google Research, Brain Team}
\icmlaffiliation{pq}{ProQuest LLC}

\icmlcorrespondingauthor{Daniel D'souza, Sara Hooker}{ddsouza@umich.edu, shooker@google.com}

\icmlkeywords{Machine Learning}
\vskip 0.3in
]

\printAffiliationsAndNotice{\icmlEqualContribution} 

\begin{abstract}
\looseness=-1
As machine learning models are increasingly employed to assist human decision-makers, it becomes critical to communicate the uncertainty associated with these model predictions.  However, the majority of work on uncertainty has focused on traditional probabilistic or ranking approaches -- where the model assigns low probabilities or scores to uncertain examples. While this captures what examples are challenging for the model, it does \emph{not} capture the underlying source of the uncertainty. In this work, we seek to identify examples the model is uncertain about \emph{and} characterize the source of said uncertainty. We explore the benefits of designing a targeted intervention – targeted data augmentation of the examples where the model is uncertain over the course of training. We investigate whether the rate of learning in the presence of additional information differs between atypical and noisy examples? Our results show that this is indeed the case, suggesting that well-designed interventions over the course of training can be an effective way to characterize and distinguish between different sources of uncertainty.
\end{abstract}

\section{Introduction}
\label{sec:intro}
As machine learning models are increasingly implemented in real-world applications, it becomes critical to understand where model predictions are uncertain and ensure that model behavior is safe and trustworthy. Traditional approaches to uncertainty estimation use a probabilistic approach -- where examples a model is uncertain about are assigned low probabilities or scores \cite{agarwal2020estimating,Baldock2021DeepLT,Denker1990,2017Hendrycks,erfani2016high,ruff2018deep,parzen1962estimation,rosenblatt1956remarks,hawkins1974detection,vandeginste1988robust}. While probabilities and other scores are an effective way to isolate a subset of examples that are high uncertainty, this estimate of uncertainty is fundamentally limited as it captures what predictions are challenging for the model but \emph{not} the underlying source of the uncertainty.

\begin{figure}[t]
	\centering
	{
	\begin{flushleft}
	    \hspace{1.4cm}{\texttt{horse}}
	    \hspace{2.6cm}{\texttt{donkey}}
	\end{flushleft}
    }
	\begin{subfigure}{0.45\columnwidth}
    	\includegraphics[width=0.9\linewidth]{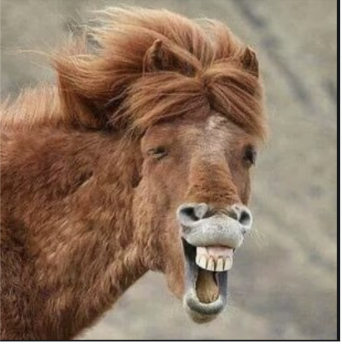}
    	\caption{Reducible Error}
        \label{fig:reducible_error}
	\end{subfigure}
	\begin{subfigure}{0.45\columnwidth}
        \includegraphics[width=0.9\linewidth]{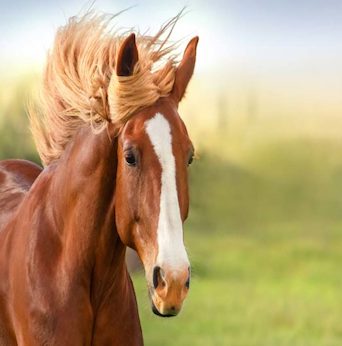}
        \caption{Irreducible Error}
        \label{fig:irreduciblerror}
	\end{subfigure}
	\caption{Examples of different predictive uncertainties. \textbf{Left:} An instance of the \texttt{horse} class representing error reducible using more data examples. \textbf{Right:} A \texttt{horse} image mislabelled as a \texttt{donkey}, representing irreducible error as the model cannot learn this class distribution even with more examples because of the corrupted label.}
	\label{fig:types_uncertainty}
\end{figure}

\looseness=-1
Most vision and language datasets exhibit a long-tail distribution with an unequal frequency of attributes in training data ~\citep{zipf1999psycho,Feldman2020}. However, the nature of these low-frequency attributes differs considerably. Atypical examples are rare or unusual attributes -- data points sampled from sparsely populated regions of the input space. Poor model performance on atypical examples reflects \emph{epistemic} uncertainty, where there is insufficient evidence for the model to learn the respective distribution. Ideally, a model spends more time learning these atypical examples than redundant high frequency attributes which are learned early during the training and can subsequently be safely ignored \citep{agarwal2020estimating,paul2021deep,Mangalam2019DoDN}.

Unlike atypical examples, noisy examples are due to influences on the data-generating process, such as label corruption or input data perturbation, which impairs the learnability of the instance. These noisy examples are dominated by \emph{aleatoric} uncertainty or \textit{irreducible error} because the mapping between the input and output space is entirely stochastic. Recent works have suggested that labeling noise is widespread in widely used datasets, and can constitute a large fraction of the training set \citep{hooker2020compressed,northcutt2021pervasive,2020arXiv200607159B}.

\begin{figure*}[ht]
	{
	\begin{flushleft}
	    \hspace{6.5cm}\textbf{CScore-Noise Dataset}
	\end{flushleft}
    }
	{
	\begin{flushleft}
	    \hspace{1.3cm}\texttt{No Augmentation}
	    \hspace{1.9cm}\texttt{Standard Augmentation}
	    \hspace{1.2cm}\texttt{Targeted Augmentation}
	\end{flushleft}
    }
    \begin{subfigure}{0.33\textwidth}
        \includegraphics[width=\columnwidth]{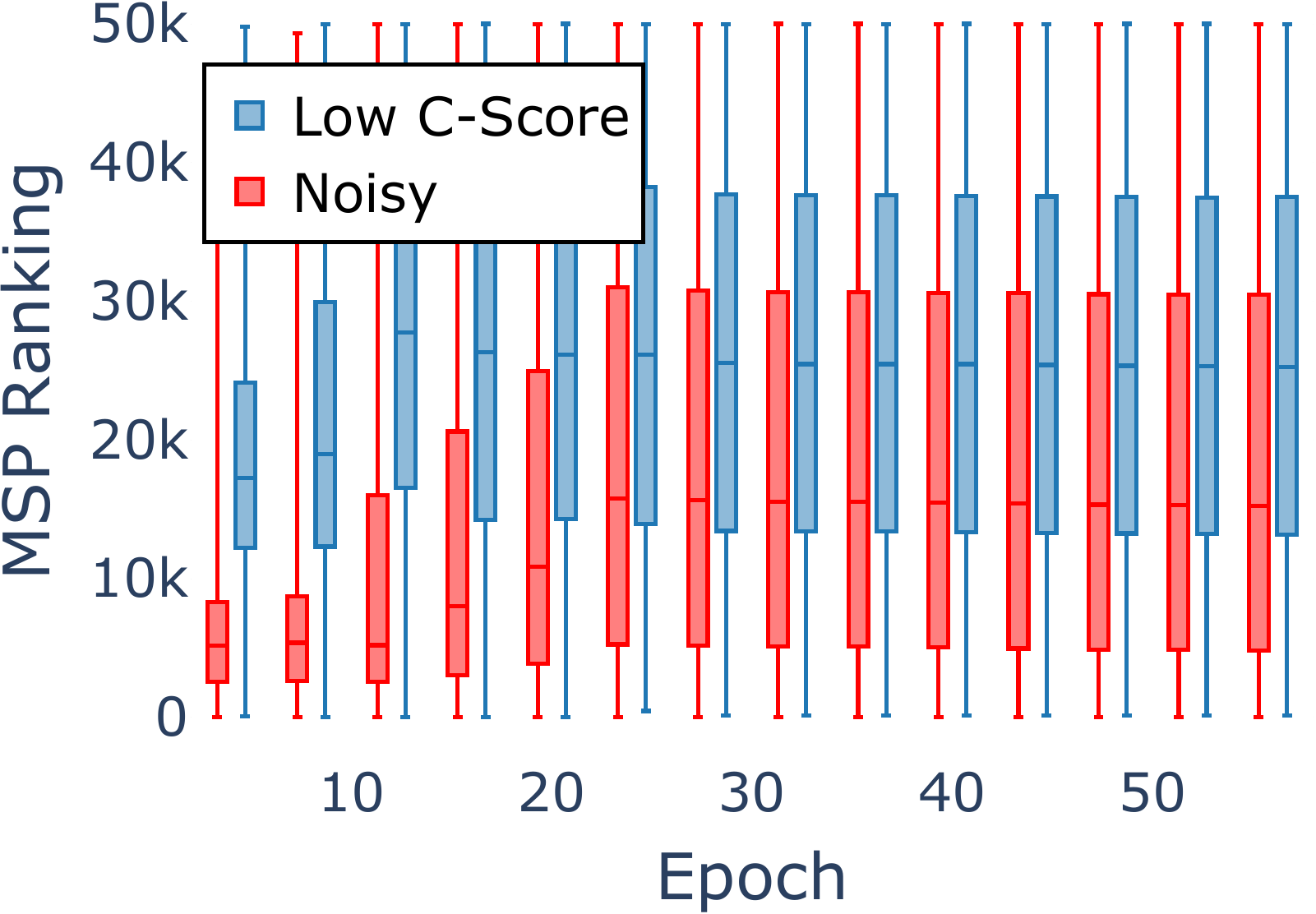}
        \label{c10-cscore-noaug-msp}
    \end{subfigure}
    \begin{subfigure}{0.33\textwidth}
        \includegraphics[width=\columnwidth]{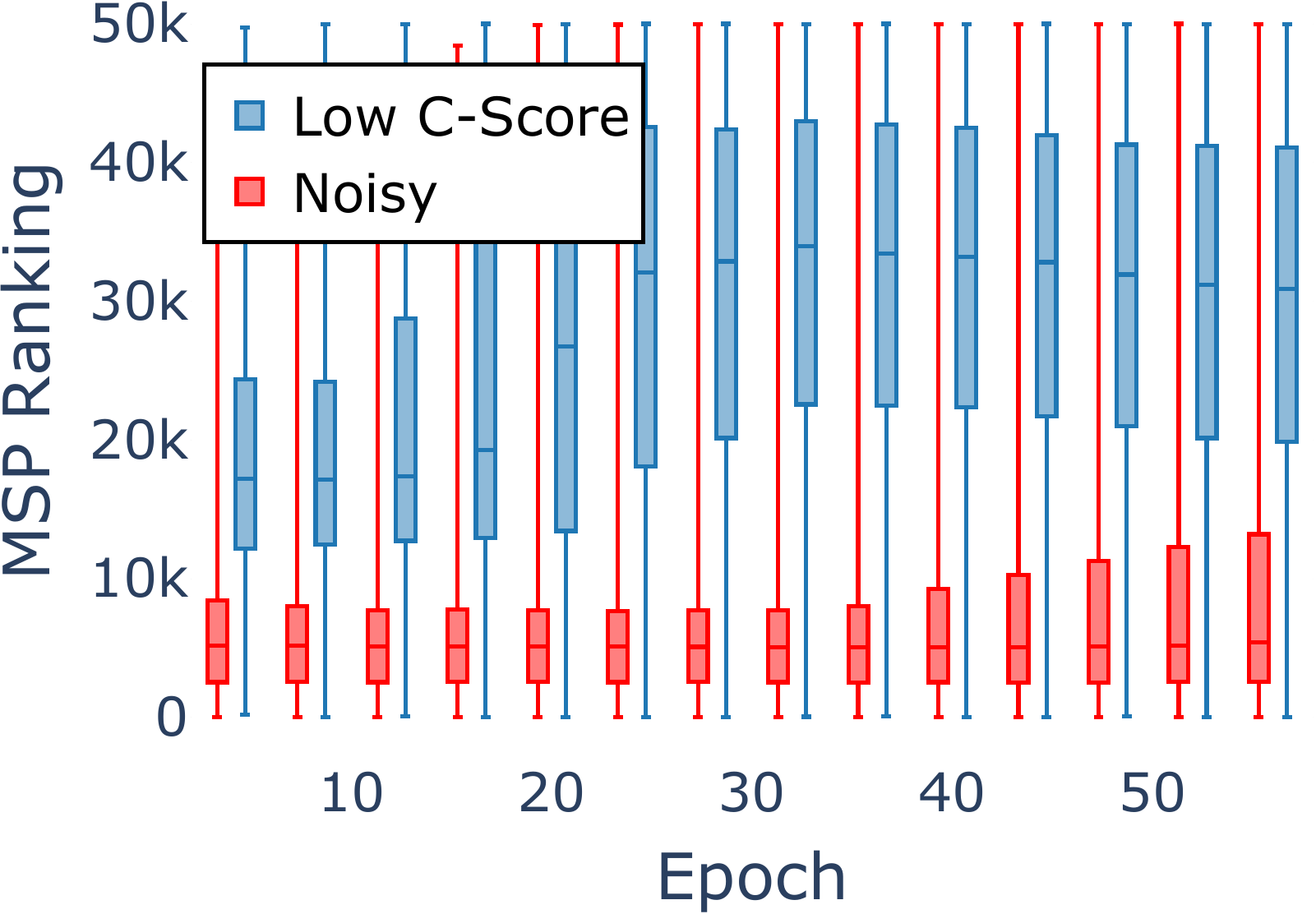}
        \label{c10-cscore-aug-msp}
    \end{subfigure}
    \begin{subfigure}{0.33\textwidth}
        \includegraphics[width=\columnwidth]{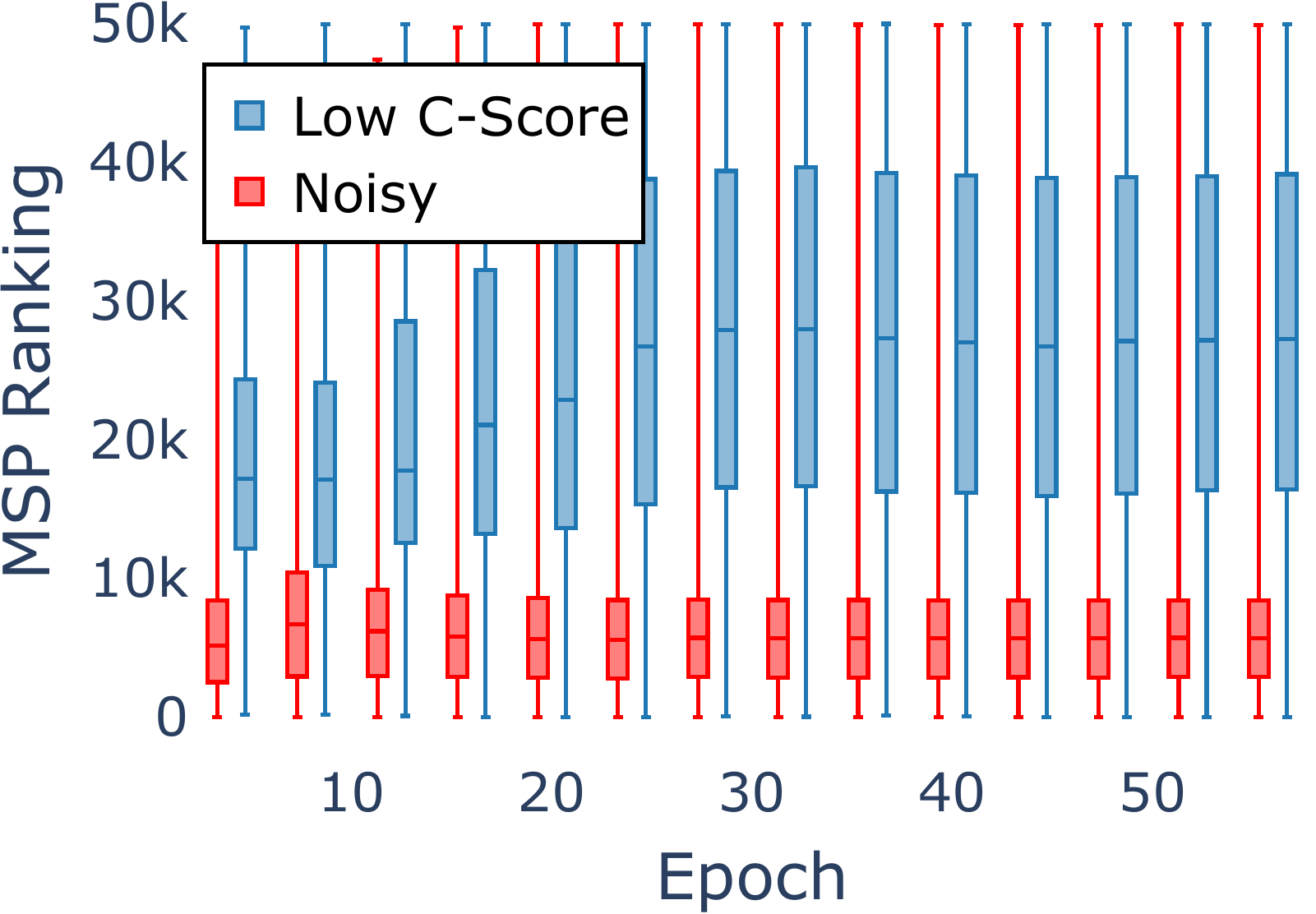}
        \label{c10-cscore-targeted-msp}
    \end{subfigure}
	{
	\begin{flushleft}
	    \hspace{6.5cm}\textbf{Frequency-Noise Dataset}
	\end{flushleft}
    }
	{
	\begin{flushleft}
	    \hspace{1.3cm}\texttt{No Augmentation}
	    \hspace{1.9cm}\texttt{Standard Augmentation}
	    \hspace{1.2cm}\texttt{Targeted Augmentation}
	\end{flushleft}
    }
    \begin{subfigure}{0.33\textwidth}
        \includegraphics[width=\columnwidth]{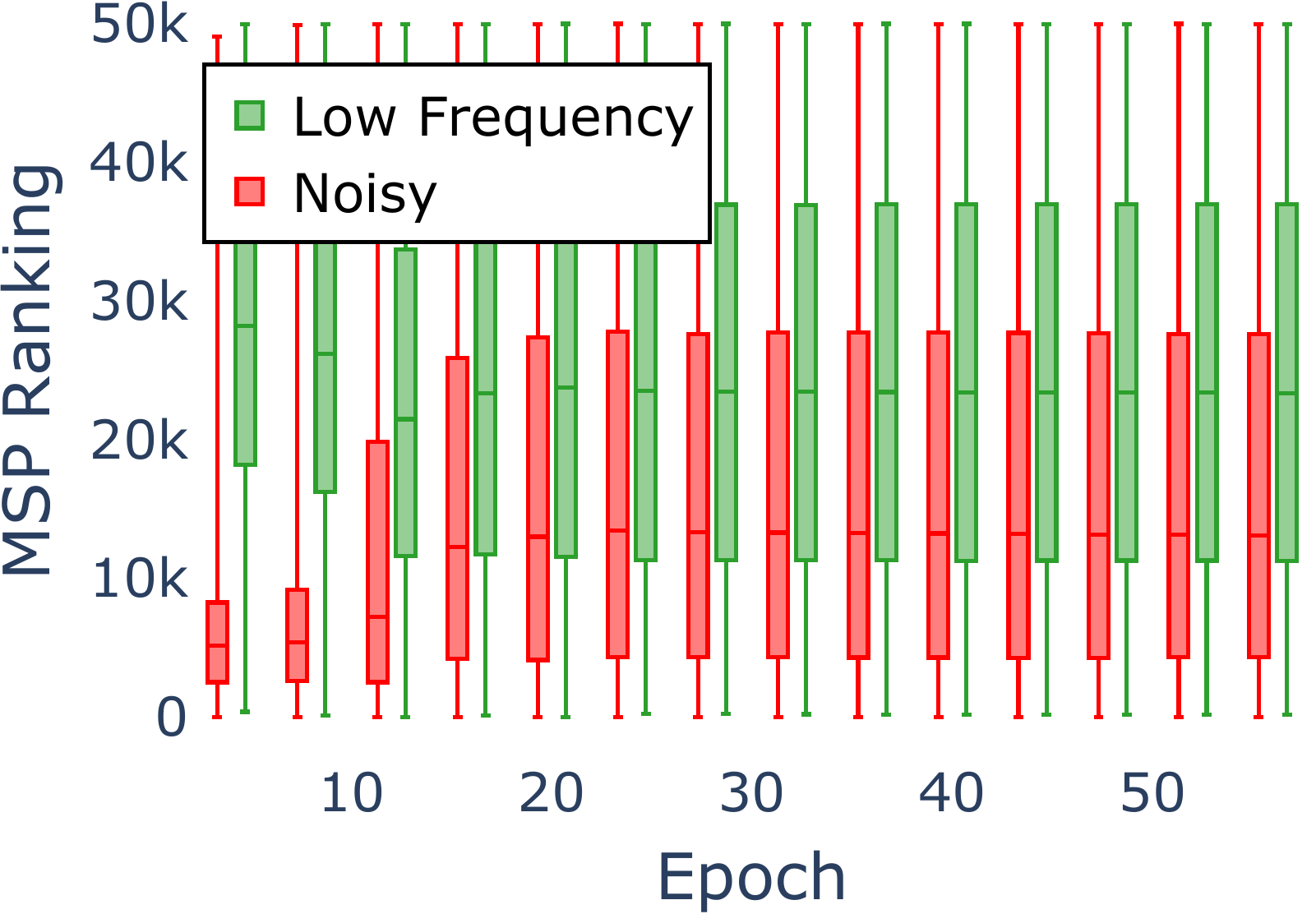}
        \label{c10-freq-noaug-msp}
    \end{subfigure}
    \begin{subfigure}{0.33\textwidth}
        \includegraphics[width=\columnwidth]{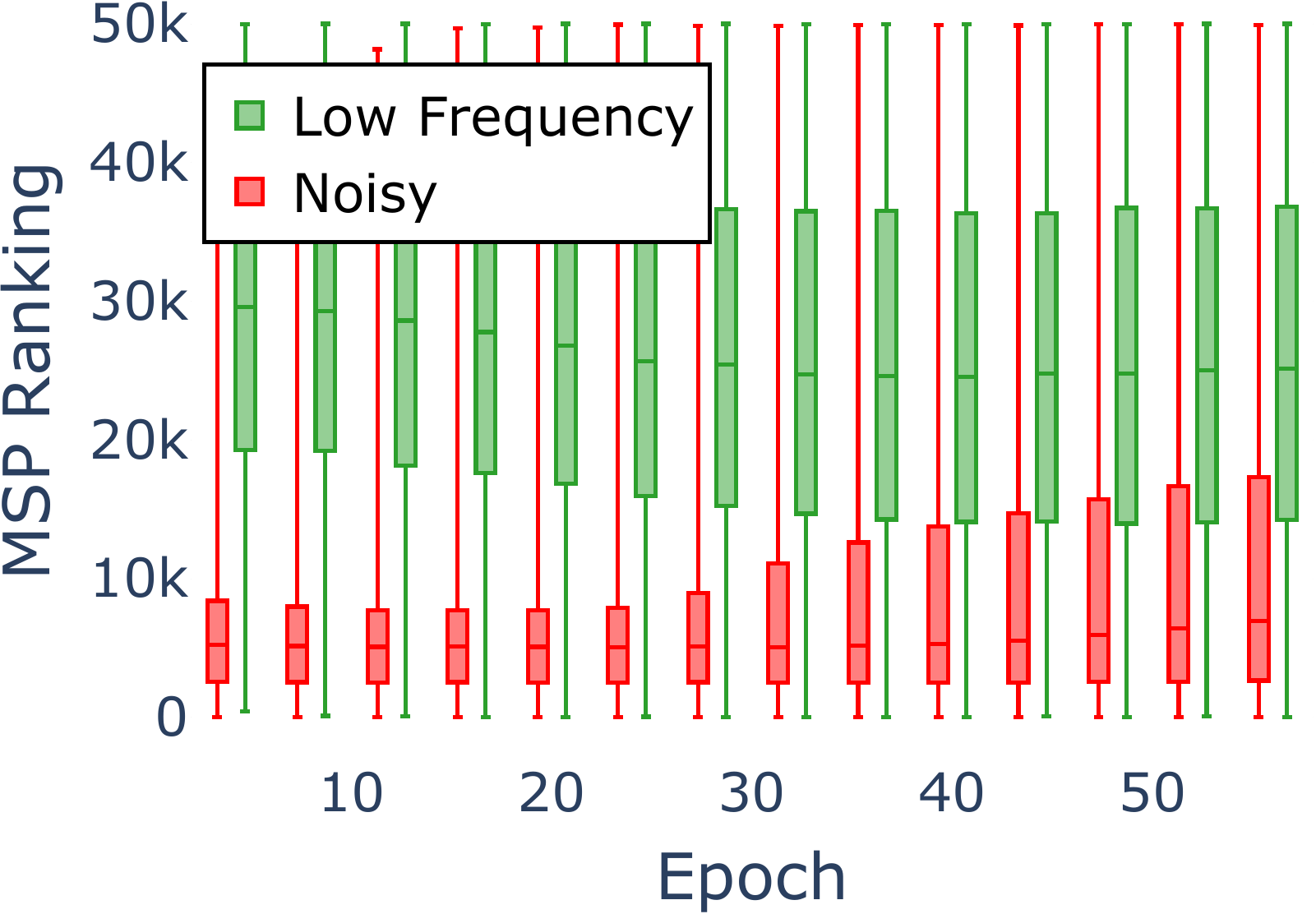}
        \label{c10-freq-aug-msp}
    \end{subfigure}
    \begin{subfigure}{0.33\textwidth}
        \includegraphics[width=\columnwidth]{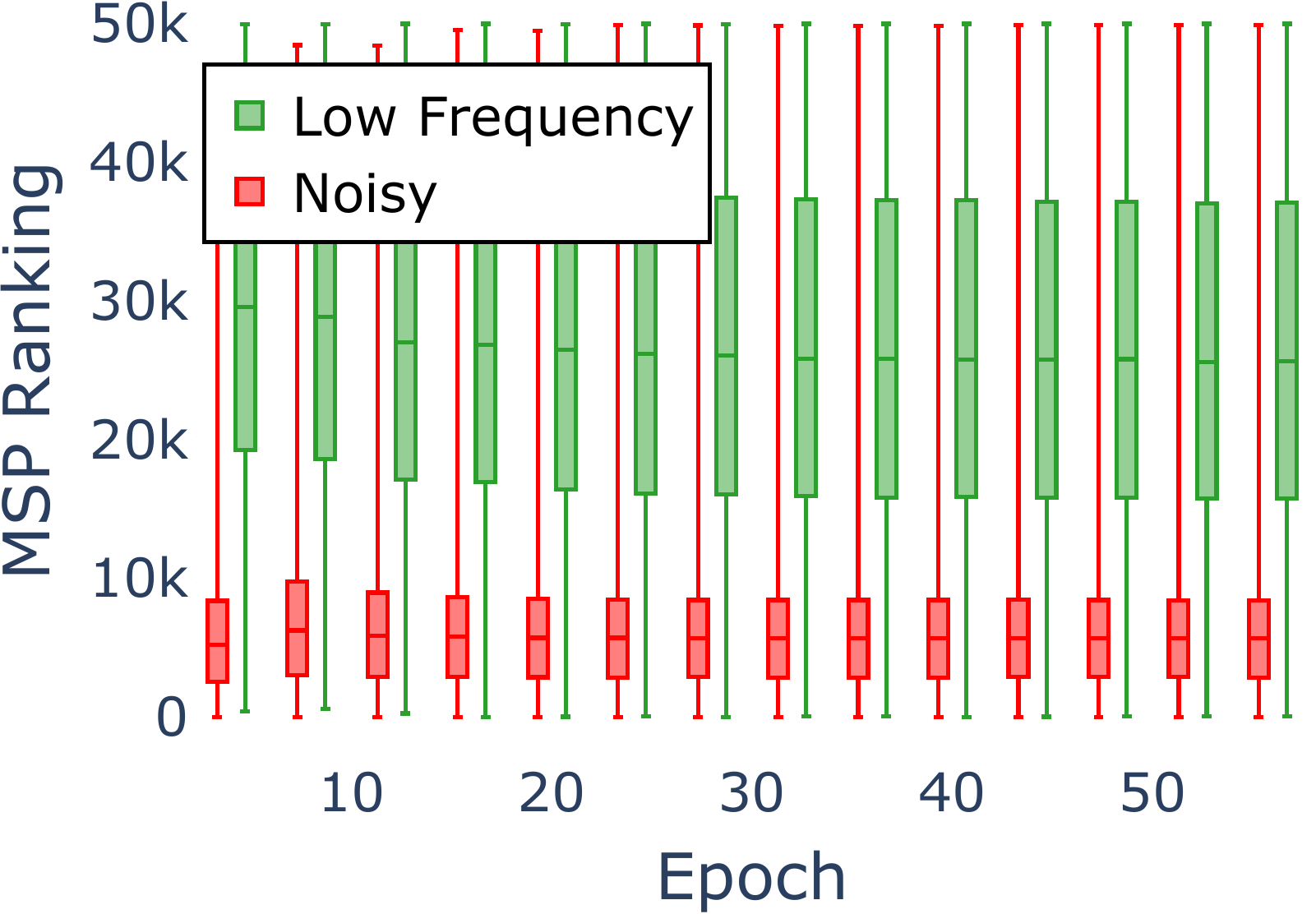}
        \label{c10-freq-targeted-msp}
    \end{subfigure}
    \vskip -0.2in
    \caption{MSP ranking for atypical and noisy subsets in LongTail Cifar-10 dataset across training for different augmentation variants}
    \label{fig:c10_aug_box_msp}
\end{figure*}

\begin{figure*}[ht]
	{
	\begin{flushleft}
	    \hspace{6.5cm}\textbf{CScore-Noise Dataset}
	\end{flushleft}
    }
	{
	\begin{flushleft}
	    \hspace{1.5cm}\texttt{No Augmentation}
	    \hspace{1.9cm}\texttt{Standard Augmentation}
	    \hspace{1.5cm}\texttt{Targeted Augmentation}
	\end{flushleft}
    }
    \begin{subfigure}{0.33\textwidth}
        \includegraphics[width=\columnwidth]{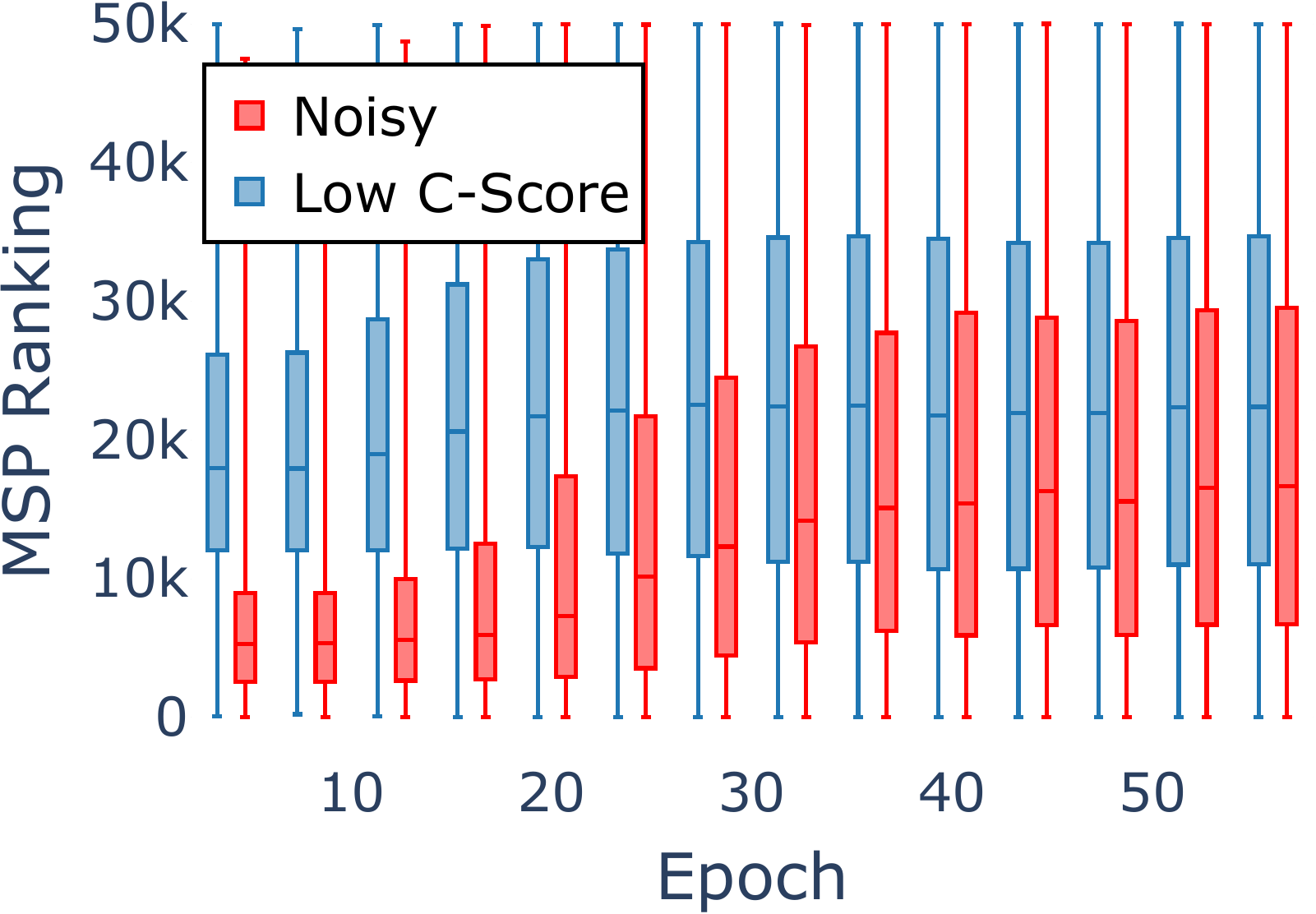}
        \label{c100-cscore-noaug-msp}
    \end{subfigure}
    \begin{subfigure}{0.33\textwidth}
        \includegraphics[width=\columnwidth]{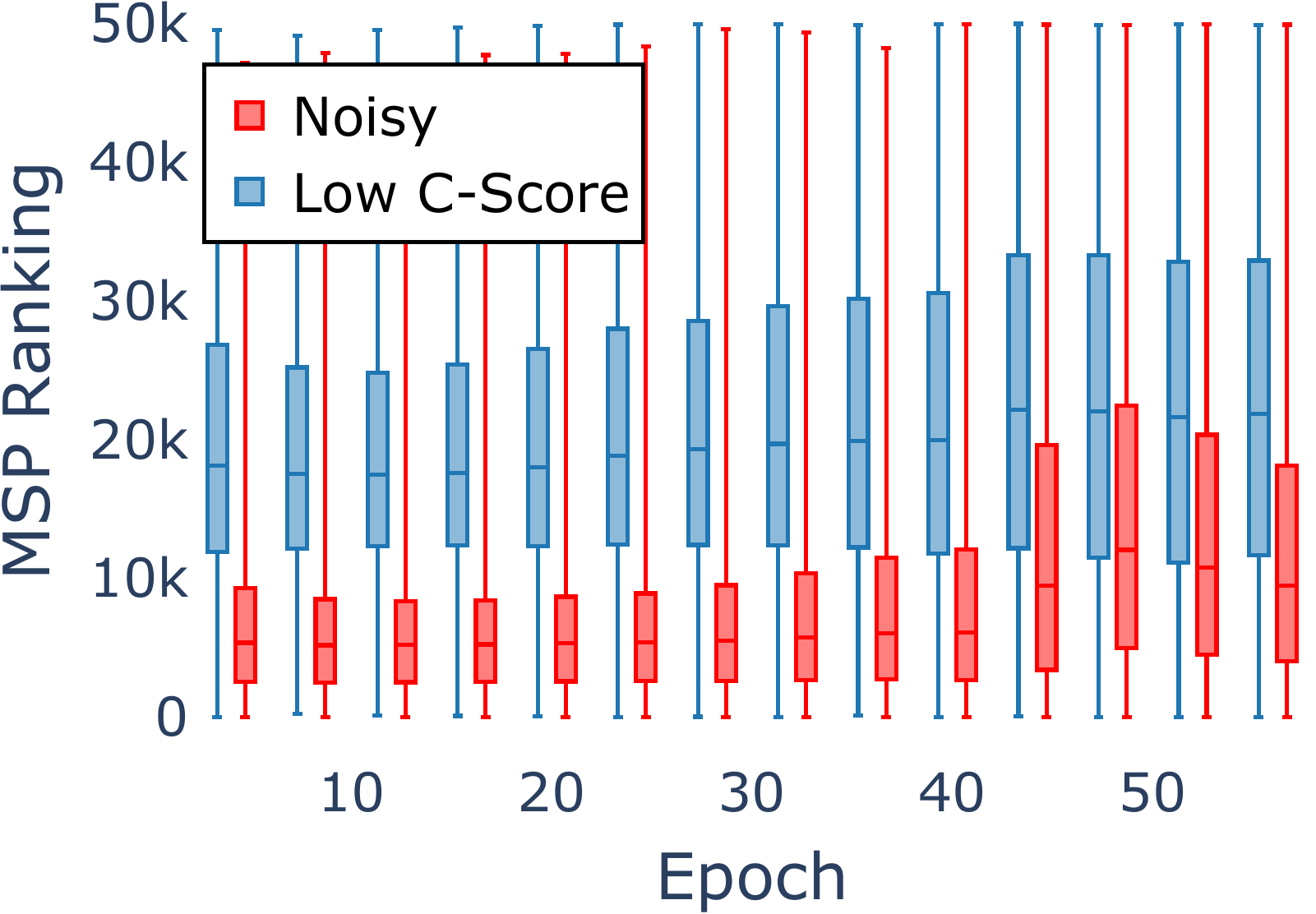}
        \label{c100-cscore-aug-msp}
    \end{subfigure}
    \begin{subfigure}{0.33\textwidth}
        \includegraphics[width=\columnwidth]{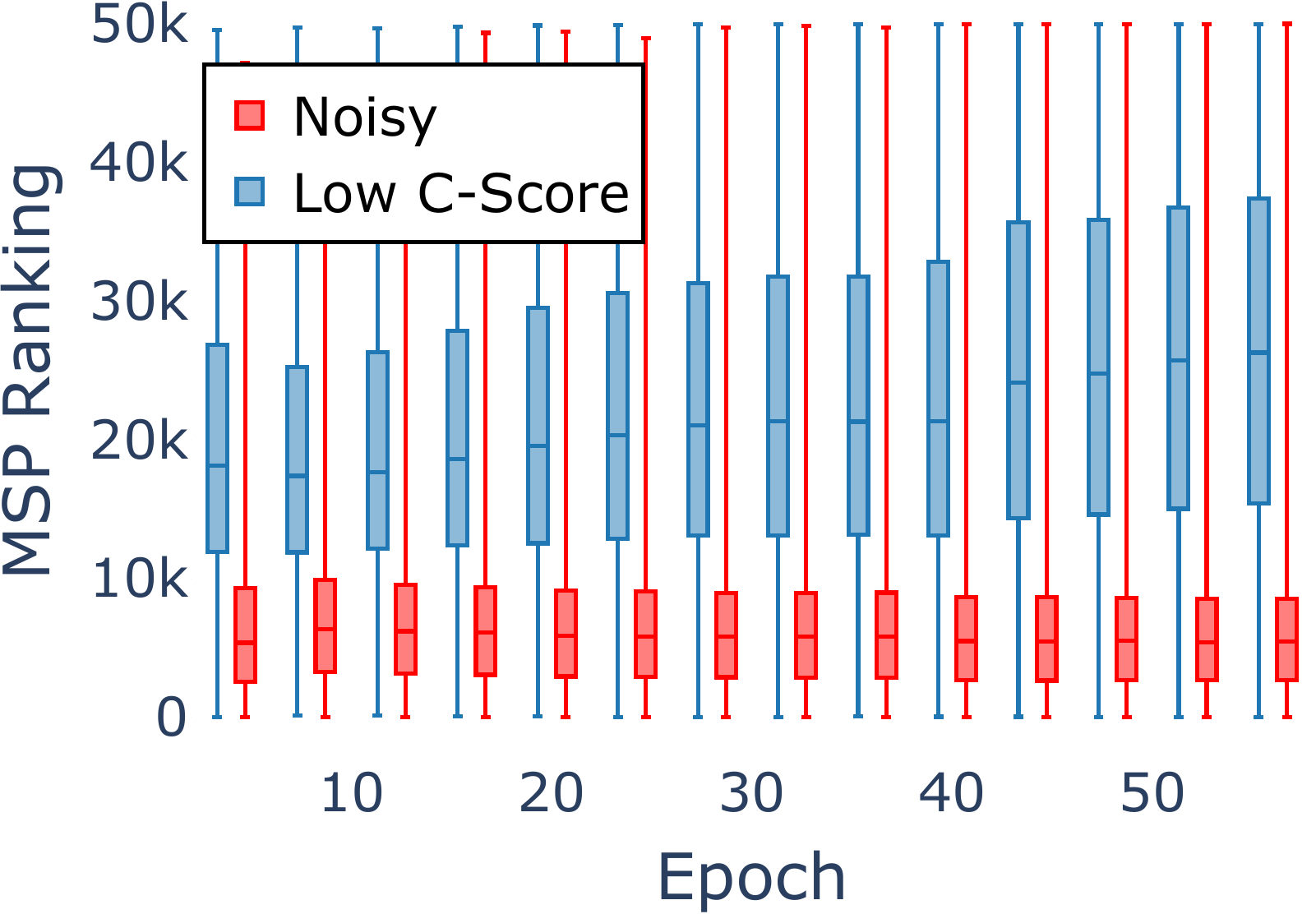}
        \label{c100-cscore-targeted-msp}
    \end{subfigure}
	{
	\begin{flushleft}
	    \hspace{6.5cm}\textbf{Frequency-Noise Dataset}
	\end{flushleft}
    }
	{
	\begin{flushleft}
	    \hspace{1.5cm}\texttt{No Augmentation}
	    \hspace{1.9cm}\texttt{Standard Augmentation}
	    \hspace{1.5cm}\texttt{Targeted Augmentation}
	\end{flushleft}
    }
    \begin{subfigure}{0.33\textwidth}
        \includegraphics[width=\columnwidth]{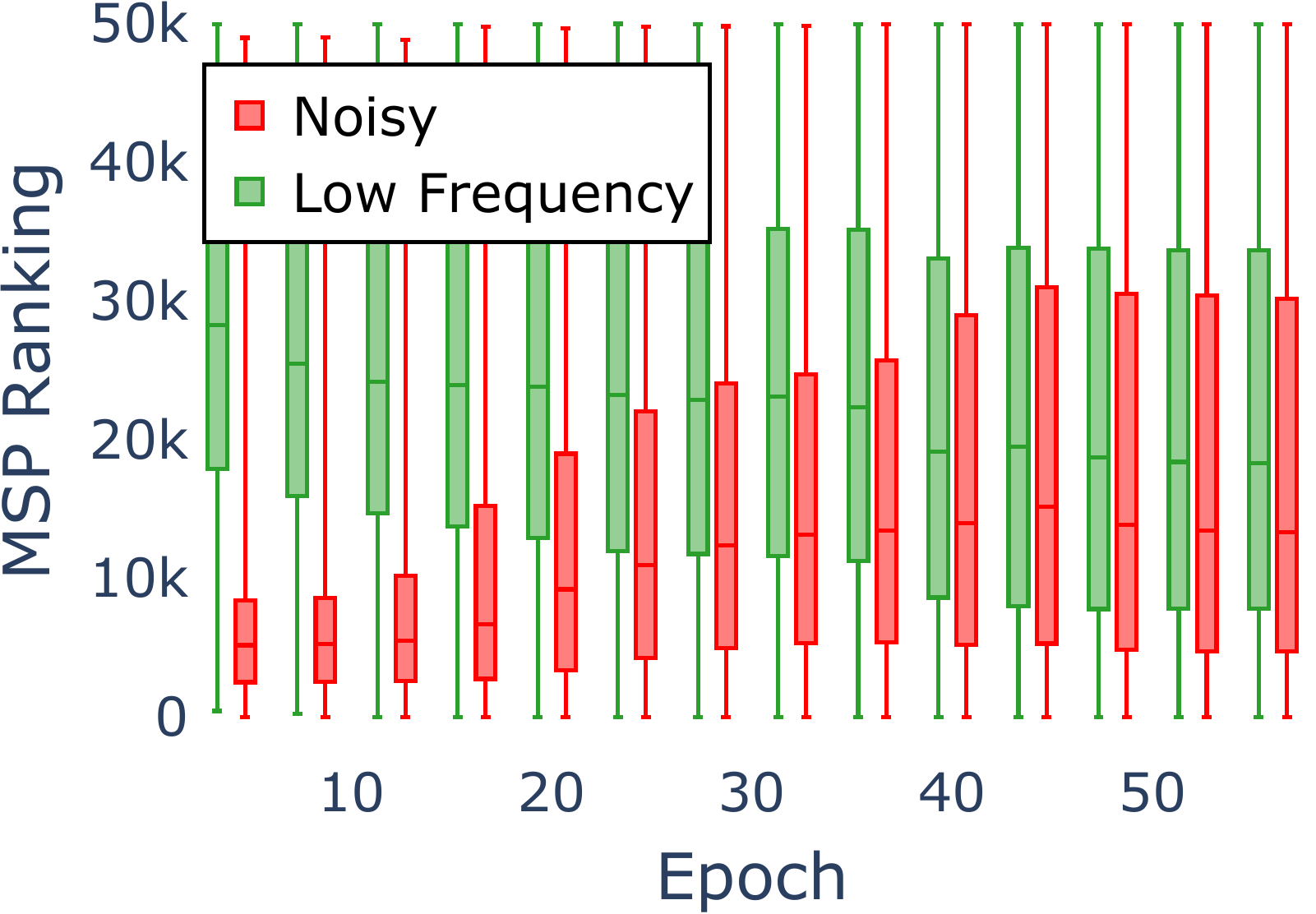}
        \label{c100-freq-noaug-msp}
    \end{subfigure}
    \begin{subfigure}{0.33\textwidth}
        \includegraphics[width=\columnwidth]{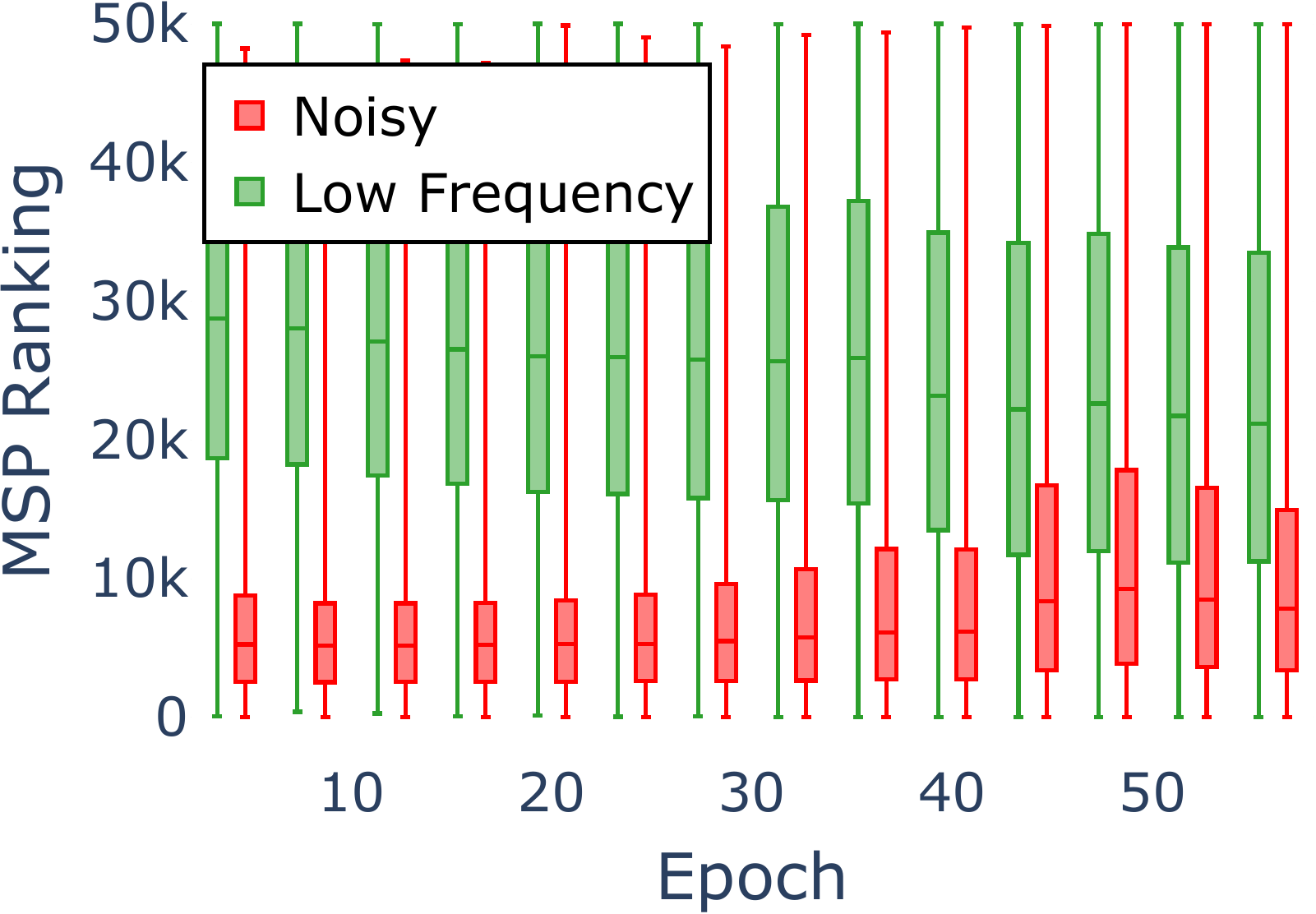}
        \label{c100-freq-aug-msp}
    \end{subfigure}
    \begin{subfigure}{0.33\textwidth}
        \includegraphics[width=\columnwidth]{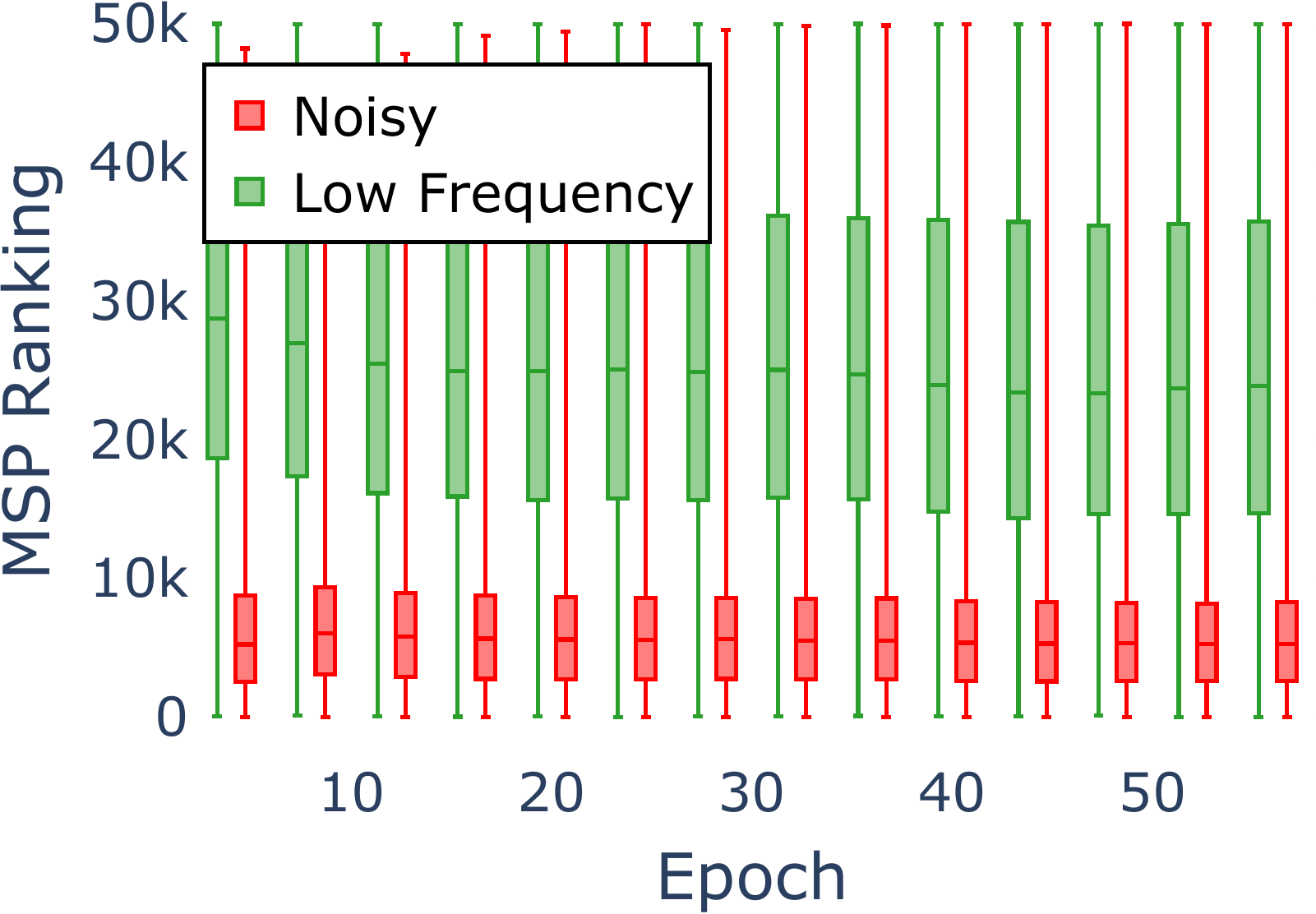}
        \label{c100-freq-targeted-msp}
    \end{subfigure}
    \vskip -0.2in
    \caption{MSP ranking for atypical and noisy subsets in LongTail Cifar-100 dataset across training for different augmentation variants}
    \label{fig:c100_aug_box_msp}
\end{figure*}

The need for a framework to estimate both the level \emph{and} source of uncertainty is driven by the very different downstream remedies for different sources of uncertainty. For sources of high \emph{epistemic} uncertainty, such as low-frequency atypical attributes or challenging fine-grained samples, a practitioner can improve model performance by either collecting more data that are similar or re-weighting examples to improve model learning of this instance (or other active learning techniques) \citep{Budd_2021,Zhang2019ToBO,Liu_2020_CVPR,4129456,shen2018deep}. In contrast, for causes of \emph{aleatoric} uncertainty, such as noisy examples, solutions like down-weighting or elimination through data cleaning are advocated \citep{Zhang_2020_CVPR,8953408,thulasidasan2019combating,Liu2020PeerLF,schroder2020survey,pleiss2020}.

\looseness=-1
Despite the importance of identifying the sources of predictive uncertainty, it has been relatively under-treated in machine learning literature. Probabilistic frameworks will accord high uncertainty to both atypical and noisy examples, failing to distinguish between the two. Moreover, optimization techniques that aim to modify training to prioritize instances with high uncertainty -- such as loss-based prioritization \citep{loshchilov2016online,kawaguchi2020ordered} and importance sampling methods \citep{katharopoulos2019samples} -- fail to distinguish between the different sources of uncertainty that dominate challenging examples. In large-scale training settings, models are often trained on datasets with unknown quality and degrees of input or label corruption \citep{hooker2020compressed,beyer2020imagenet,Tsipras2020FromIT,dehghani2021benchmark}. Recent work empirically demonstrates that loss-based acceleration methods degrade in scenarios with even a limited amount of noisy and corrupted instances in the dataset \citep{hu2021does,paul2021deep}. While sub-fields have evolved separately around the treatment of low-frequency \citep{hooker2020characterising,pmlr-v81-buolamwini18a,hashimoto18a,Sowik2021AlgorithmicBA} and noisy distributions \citep{wu2020topological,yi2019probabilistic,pmlr-v97-thulasidasan19a}, only limited work to date has focused on the sources of uncertainty within a unified framework \citep{NIPS2017_2650d608,pmlr-v80-depeweg18a}.

\looseness=-1
\xhdr{Present Work} In this work, we seek to identify examples the model is uncertain about \emph{and} characterize the source of said uncertainty. We leverage the key difference between \emph{Epistemic} and \emph{Aleatoric} uncertainty -- one error is reducible in the presence of additional data, and the other is not. We propose targeted data augmentation throughout training to amplify the difference in learning rate between atypical and noisy examples. Our results show well-designed interventions throughout training can be an effective way to cluster and distinguish between different sources of uncertainty.

\section{Methodology}
\label{sec:method}
\subsection{Sources of Uncertainty}
\label{sec:uncertain}
We consider a supervised learning setting where we denote the training dataset $\mathcal{D}$ as:
\begin{equation}
\mathcal{D} \defeq \big\{(x_{1}, y_{1}),\dots ,(x_{N}, y_{N})\big\} \subset \mathcal{X} \times \mathcal{Y} \enspace,
\end{equation}
where $\mathcal{X}$ represents the data space and $\mathcal{Y}$ the set of outcomes associated with the respective instances. We consider a neural network as a function $f_{w}: \mathcal{X} \mapsto \mathcal{Y}$ with trainable weights $w$. Given the training dataset, $f_{w}$ optimizes a set of weights $w^*$ by minimizing an objective function $L$,
\begin{equation}
w^\ast = \argmin_w L(w)
\end{equation}
\looseness=-1
Here, we aim to quantify the uncertainty associated with a model prediction and to subsequently identify the source of the uncertainty by classifying examples contributing disproportionately to \textit{aleatoric} or \textit{epistemic} uncertainty. Thus, we firstly would like to obtain a good measure of \emph{predictive uncertainty} related to the prediction $\hat{y}_{i}$ for an input instance $\mathbf{x}_{i} \in \mathcal{X}$.

An accurate estimate of predictive uncertainty $s(f_w,\mathbf{x}_i)$ for an instance $\mathbf{x}_{i} \in \mathcal{X}$ will reflect the accumulation of uncertainty in the data curation process, the set of modeling choices, and the training protocol itself. Thus, the possible outcome $\hat{y}_{i}$ depends upon the dataset $\mathcal{D}$ and the underlying model $f_{w}$. Intuitively,  $s(f_w,\mathbf{x}_{i})$ is a composition of both aleatoric ($s_{a}(f_{w},\mathbf{x}_{i})$) and epistemic knowledge ($s_{e}(f_{w},\mathbf{x}_{i})$).


In this work, we leverage Maximum Softmax Probability (MSP) \citep{2017Hendrycks} as our estimate of overall example uncertainty. MSP is considered state-of-the-art at classifying datapoints as being out-of-distribution \citep{2016Hendrycks,2020arXiv200406100H} and recent work has found it be a valuable proxy for model uncertainty \citep{minderer2021revisiting} despite the simplicitly in formulation. While MSP is typically computed as the predicted softmax probability at test time \citep{2017Hendrycks}, our focus is on training dynamics so we use the softmax probability for the true label


\subsection{Distinguishing Between Aleatoric and Epistemic Uncertainty}

One way to characterize uncertainty as aleatoric or epistemic is to ask whether it can be reduced through additional training data. In this work, we apply transformations $\mathcal{T}$ to the training set $\mathcal{D}$, resulting in a new set $\mathcal{D}_A$. The stochasticity of the transformation parameters is responsible for generating new samples, \ie \textit{data augmentation}. 

We evaluate the impact of providing additional information for all training examples $\forall \mathbf{x}_{i} \in \mathcal{X}$ relative to not providing additional information. This amounts to comparing data augmentation uniformally applied to all examples to no data augmentation. In addition, we also explore the benefits of designing a targeted intervention -- augmentation selectively applied to samples which are estimated to be high uncertainty. For the first 3 epochs, targeted augments 100\% i.e all of the samples. From the $4^{\text{th}}$ epoch, we utilize the MSP score to rank and selectively augment the bottom 20\% MSP percentile at every epoch for the rest of training. 

Given our hypothesis that additional information will help distinguish between reducible and irreducible error, we expect differences in the distributions of atypical and noisy to be amplified for the two augmentation schemes (\texttt{standard augmentation} and \texttt{targeted augmentation}) \emph{relative} to no augmentation. Given our assumption is that additional information is most valuable where the model is uncertain, we also seek to understand if \texttt{targeted augmentation} can have comparable impact to \texttt{standard augmentation} despite being applied to a fraction of the total dataset.

\subsection{Experimental Framework} \label{subsec:experimental_framework}

To understand how atypical and noisy examples are learned, we construct atypical and noisy subsets of the training set where ground truth is known for these examples. We briefly describe the subset construction below:

\xhdr{Atypical sub-sets} We construct two different atypical sub-sets based upon two different notions of typicality: 1) \emph{Frequency} and 2) \emph{Consistency}.
\begin{enumerate}
\item {\bf} \textbf{Frequency.~} We artificially create a known frequency disparity between examples. We sample a uniform fraction $p$ at random from each class in the training set (hence preserving class balance). Of the remaining dataset, we sample a fraction $t$ and create two copies of each example, where $t$ is selected such that the overall dataset size is the same as the original unmodified dataset.
\looseness=-1
\item {\bf} \textbf{Memorized examples.~} We utilize the C-Score \citep{2020Jiang} -- a pseudo-measure of how \emph{typical} an example is \textit{wrt} other samples in the dataset. Here, a low score indicates a highly atypical example that is inconsistent with the wider data corpus. We directly use the pre-computed C-scores \footnote{Available from \url{https://pluskid.github.io/structural-regularity/}} as a continuous measure of (a)typicality.

\end{enumerate}
\xhdr{Noisy} 
\looseness=-1
In addition to the two atypical subsets, we also craft a noisy ground truth. For modeling a noisy subset, we follow \citet{zhang2016understanding} and assign uniformly shuffled labels to a percentage of the training data. More specifically, this decision models noisy data as mislabelled instances.

\xhdr{Dataset Construction} We construct two different datasets\footnote{Code/Datasets available at : \url{https://github.com/dsouzadaniel/long_tail}} \texttt{Frequency-Noise Dataset} and \texttt{CScore-Noise Dataset} based upon the subsets described above for Cifar-10 and Cifar-100 \cite{krizhevsky2009learning}. These datasets differ by the choice of the atypical subset. For both variants, we maintain a ratio of 20\% noisy, 20\% atypical, and 60\% typical examples and ensure that it is the same size as the original dataset.

\looseness=-1

\begin{enumerate}
\item {\bf} \textbf{Frequency-Noise Dataset} We uniformly sample 20\%   training set as the low-frequency atypical subset. We subsequently shuffle the labels of a 20\% random selection from the remaining examples. Finally, we sample 30\% from the remaining dataset, create two copies and add that as typical candidates. We note that although the dataset size remains the same as the baseline, this variant does not use $30 \%$ of the original dataset. 

\item {\bf} \textbf{CScore-Noise Dataset} For the \texttt{CScore-Noise} dataset, we consider the bottom 20\% C-Score ranked images as atypical.
Similar to the \texttt{Frequency-Noise} dataset, we then sample another random 20\% from the remaining examples, uniformly shuffle the labels and add these as noisy candidates. The remaining 60\% of the dataset is considered as typical. 
\end{enumerate}

We note that \texttt{CScore-Noise} preserves all original datapoints in the Cifar-10/Cifar-100 training set, whereas \texttt{Frequency-Noise} downsamples the number of original examples in order to maintain the same training set size. While \texttt{Frequency-Noise} and \texttt{CScore-Noise} differ in the construction of the atypical subset, both have the same fraction of noisy examples.

\xhdr{Training details} We conduct experiments on both Cifar-10 and Cifar-100. For all our Cifar experiments, we use a WideResNet \citep{Zagoruyko2016} architecture. We train for $60$ epochs using stochastic gradient descent (SGD). For training variants where augmentation is present, we use standard data augmentation by applying random horizontal flips and crops with padding. We use a base learning rate schedule of $0.1$ and adaptively dampen it by a factor of $0.2$ at the $10^{\text{th}}$, $20^{\text{th}}$ and $30^{\text{th}}$ training epochs for Cifar-10 and at the $40^{\text{th}}$, $50^{\text{th}}$ and $55^{\text{th}}$ training epochs for Cifar-100. For our baseline variant on a clean dataset (no artificial stratification of noisy and atypical) we report a top-1 test-set accuracy of $93.11\%$ and $77.46\%$ for Cifar-10 and Cifar-100 respectively.

\begin{table}[t]
\setlength{\tabcolsep}{3.0pt}
\caption{Testing accuracy given different augmentation variants of LongTail Cifar datasets.}
\label{frequency-metrics-table}
\begin{center}
\begin{small}
\begin{sc}
\begin{tabular}{llcrc}
\toprule
\multirow{2}{2cm}{Dataset} & \multirow{2}{2cm}{Variant} & \multicolumn{2}{c}{Test Accuracy} \\
 &  & Cifar-10 & Cifar-100\\
\midrule
\multirow{3}{2cm}{Frequency} & No Aug. &  60.7\% & 36.7\%\\
& Standard Aug. &  67.7\% & 54.3\% \\
& Targeted Aug. &  83.1\% & 48.8\%\\
\midrule
\multirow{3}{2cm}{C-Score} & No Aug. &  72.1\% & 42.7\% \\
& Standard Aug. & 77.0\% & 60.4\% \\
& Targeted Aug. & 85.4\% & 56.4\%\\
\bottomrule
\end{tabular}
\end{sc}
\end{small}
\end{center}
\vskip -0.15in
\end{table}

\section{Results}\label{sec:results}


In this section, we address a key question: \textit{Can the presence of additional information amplify differences in the rate of learning of atypical and noisy examples?} We note that we observe consistent results across both datasets \texttt{Frequency-Noise Dataset} and \texttt{CScore-Noise Dataset} for both Cifar-10 and Cifar-100.

\xhdr{Characterizing differences between atypical and noisy subsets across training} In Fig.~\ref{fig:c10_aug_box_msp} and Fig.~\ref{fig:c100_aug_box_msp}, we plot the distribution of ranks based on MSP scores for both noisy and atypical samples across training. We now describe the effect of different augmentation variants on separating the noisy and atypical subsets.

\xhdr{a) No Augmentation} In both Fig.~\ref{fig:c10_aug_box_msp} and Fig.~\ref{fig:c100_aug_box_msp}, we observe a large overlap in the distribution of atypical and noisy examples in a training setting without augmentation. This overlap becomes progressively more pronounced over the course of training -- as both atypical and noisy examples are accorded the same range of low probabilities. By the end of training -- it is not possible to distinguish different sources of uncertainty using a probabilistic ranking.

\xhdr{b) Standard Augmentation} We observe far better separation between the noisy and atypical distributions using standard augmentation during training (Fig.~\ref{fig:c10_aug_box_msp} and Fig.~\ref{fig:c100_aug_box_msp}). The addition of information, even if done uniformly, provides the model with additional examples of the atypical instances. Atypical MSP ranks rise more markedly, while noisy MSP ranks remain at the bottom of the MSP probability distribution.  Of interest, is that towards the end of training noisy MSP scores begin to rise steadily causing a small degree of overlap in the distributions. We attribute this to the propensity for memorization of noisy examples in the latter stages of training. 

\xhdr{c) Targeted Augmentation}  In Fig.~\ref{fig:c10_aug_box_msp} and Fig.~\ref{fig:c100_aug_box_msp}, it is clear that targeted augmentation provides a notable improvement over \emph{both} \texttt{no augmentation} and \texttt{standard augmentation}. \texttt{Targeted augmentation} is the most effective intervention strategy that distinguishes between atypical and noisy example in both datasets. Unlike \texttt{standard augmentation}, targeted augmentation preserves separation of these distributions for the entirety of training. When addition of information is specifically targeted toward the uncertain examples, it provides additional opportunities to learn atypical examples whilst appearing to better prevent memorization of noisy examples.

\section{Related Work}

Our work explores leveraging early training signals and targeted interventions to distinguish between sources of uncertainty. This differs from work which is concerned with simply estimating the overall level of per-example uncertainty (ranking example difficulty or influence) or surfacing prototypical examples (where is uncertainty is low and there is minimal redundancy of features in the dataset).

\xhdr{Ranking example difficulty or influence} \citet{2017koh} proposes influence functions to identify training points most influential on a given prediction. Work by \citep{arpit2017closer} and \citep{li2020gradient} develop methods that measure the degree of memorization required of individual examples. While \citet{2020Jiang} proposes a C-score to rank each example by alignment with the training instances, \citet{2019carlini} considers several different measures to isolate prototypes that could conceivably be extended to rank the entire dataset. \citep{agarwal2020estimating} leverage variance of gradients across training to rank examples by learning difficulty. Further, \citet{2019arXiv191105248H} classify examples as outliers according to sensitivity to varying model capacity. These methods are unified by the goal of identifying examples that the model finds challenging, but unlike our work, do not distinguish between the sources of uncertainty.

\xhdr{Core-set selection techniques} These methods aim to find prototypical examples that represent a larger corpus of data. Previous work \citep{Zhang1992,2012Bien,2015kim,NIPS2016_6300} introduces the notion of prototypes, quintessential examples in the dataset, but do not focus on deep neural networks. Work has also explored using the identification of a core set of representative prototypes to speed up training \citep{sener2018active,Shim2021CoresetSF,huggins2017coresets} or for interpretability purposes \citep{yoon2019rllim}. Recent work has focused on using a hold-out validation set to isolate a subset of the training data to train that is most conducive for generalization \citep{killamsetty2021glister,paul2021deep}. Unlike our work, core-set selection techniques aim to reduce data redundancy by eliminating unnecessary data points. In contrast, our work focuses on distinguishing different sources of uncertainty in the long-tail of the distribution.

\xhdr{Early stages of training dynamics} In this work, we leverage early stage training dynamics to create targeted interventions. Recent work has shown that there are distinct stages to training in deep neural networks \citep{Achille2017CriticalLP, jiang2020exploring, Mangalam2019DoDN, 2020fartash} and that stages of training can be used to identify challenging examples \citep{agarwal2020estimating}.

\section{Conclusion}

We leverage targeted augmentation interventions to characterize examples as dominated by \emph{aleatoric} and \emph{epistemic} uncertainty. We empirically show how augmentation protocols (both targeted and standard) amplify the differences in distribution between noisy and atypical examples. We observe that the rate of learning in the presence of additional information can be used to distinguish between atypical and noisy examples. Targeted addition of information \emph{outperforms} both \texttt{no augmentation} and \texttt{standard augmentation} at successfully clustering noisy and atypical examples.

Our results suggest that targeted interventions are a powerful tool to characterize and distinguish between different sources of uncertainty. This opens up future research directions in appropriately treating atypical and noisy subsets in a dataset during training. Common remedies for noisy examples include data cleaning, isolating a subset for re-labeling, whereas atypical examples may benefit from data augmentation, re-weighting or additional data collection. Our work motivates a more nuanced treatment of uncertainty estimation, and further development of estimators that ascribe both a level of uncertainty as well as characterization of the source of said uncertainty. 

\nocite{langley00}
\bibliography{main}

\end{document}